\documentclass[english]{article}
\usepackage{lmodern}

\usepackage[T1]{fontenc}
\usepackage[latin9]{inputenc}
\usepackage{graphicx}
\usepackage{nips15submit_e,times}

\makeatletter
\@ifundefined{showcaptionsetup}{}{%
 \PassOptionsToPackage{caption=false}{subfig}}
\usepackage{subfig}
\makeatother

\usepackage{babel}
\begin{document}

\title{Hierarchical Latent Word Clustering}

\author{
Halid Ziya Yerebakan \\
Computer and Information Science\\
IUPUI\\
Indianapolis, IN 46202\\
\texttt{hzyereba@cs.iupu.edu} \\
\And
Fitsum Reda \\
SIEMENS Healthcare \\
Malvern, PA \\
\texttt{fitsum.reda@siemens.com} \\
\AND
Yiqiang Zhan \\
SIEMENS Healthcare \\
Malvern, PA \\
\texttt{yiqiang.zhan@siemens.com} \\
\And
Yoshihisa Shinagawa\\
SIEMENS Healthcare \\
Malvern, PA \\
\texttt{yoshihisa.shinagawa@siemens.com} \\
}

\newcommand{\fix}{\marginpar{FIX}}
\newcommand{\new}{\marginpar{NEW}}

\nipsfinalcopy 

\maketitle
\begin{abstract}This paper presents a new Bayesian non-parametric model by 
extending the usage of Hierarchical Latent Dirichlet Allocation to extract tree structured word clusters from text data. 
The inference algorithm of the model collects words in a cluster if they share similar distribution over documents. 
In our experiments, we observed meaningful hierarchical structures on NIPS corpus and radiology 
reports collected from public repositories.\end{abstract}

\section{Introduction}

Extracting relationships between words plays an important role on information
retrieval, document summarization, and document classification. 
One way to extract hidden relationships is creating clusters on words.
Creating hierarchy within these clusters is further useful to enhance abstraction level.
For example, the terms \textquotedblleft brain\textquotedblright{}
and \textquotedblleft skull\textquotedblright{} are more related
as compared to \textquotedblleft brain\textquotedblright{} and \textquotedblleft kidney\textquotedblright{}
in the medical domain. We have discovered that modifying the data generation model of Hierarchical Latent
Dirichlet Allocation (HLDA) \cite{Blei04hierarchicaltopic} creates a model that is able to capture  
hidden hierarchical word clusters from text. We named the algorithm as "Hierarchical Latent Word Clustering''
(HLWC). HLWC merges generative topic models with entropy-based agglomerative word clustering methods. 

Topic modeling literature models word relationships using multinomial distributions on words. Documents are represented as mixtures of extracted topics. Latent Dirichlet Allocation (LDA) \cite{blei2003latent} is an important milestone in this area. LDA uses a Dirichlet prior on Probabilistic Latent Semantic Indexing (PLSI) \cite{hofmann1999probabilistic} to avoid overfitting.  Later, Hierarchical Dirichlet Process (HDP) \cite{teh2006hierarchical}
is extended from LDA to achieve a non-parametric structure. HDP allows to grow number of topics with incoming data. However, in these models interesting hierarchical  relations are
not captured. Extension studies considered hierarchical
versions of topic models. Authors of \cite{Blei04hierarchicaltopic}
proposed HLDA model to get tree structured topic relations from a corpus. Moreover, Pachinko Allocation Machine (PAM) allows DAG-structured hierarchy \cite{li2006pachinko}. As opposed to these models, we utilized multinomial on documents. As a result, HLWC creates word clusters instead of multinomial distributions on words. 

An alternative to get clustering of words is representing words in an embedding space 
as in Word2Vec \cite{mikolov2013efficient}, then obtaining clustering with standard clustering algorithms. However, hierarchical extraction
is a more challenging problem. Recently, REBUS algorithm is introduced
using the projection entropy value to obtain agglomerative hierarchical
clustering on words. This algorithm merges clusters pairwise to obtain final hierarchy. However, extracted relationships do not define abstract representations of  
documents because of the unbalanced binary tree structure.
Liu et. al. \cite{liu2014hierarchical} presented another greedy agglomerative procedure to create word clusters based on mutual information. In their study, a node can have multiple
branches. Our study differentiates from agglomerative methods by using
Bayesian formalism on clustering. A top-down approach divides words into 
clusters with a Bayesian non-parametric prior.

\begin{figure}
\begin{centering}
\begin{minipage}[t]{0.5\columnwidth}%
\begin{center}
\subfloat[HLWC model relations\label{fig:HLC-model-relations}]{\protect\begin{centering}
\protect\includegraphics[scale=0.37]{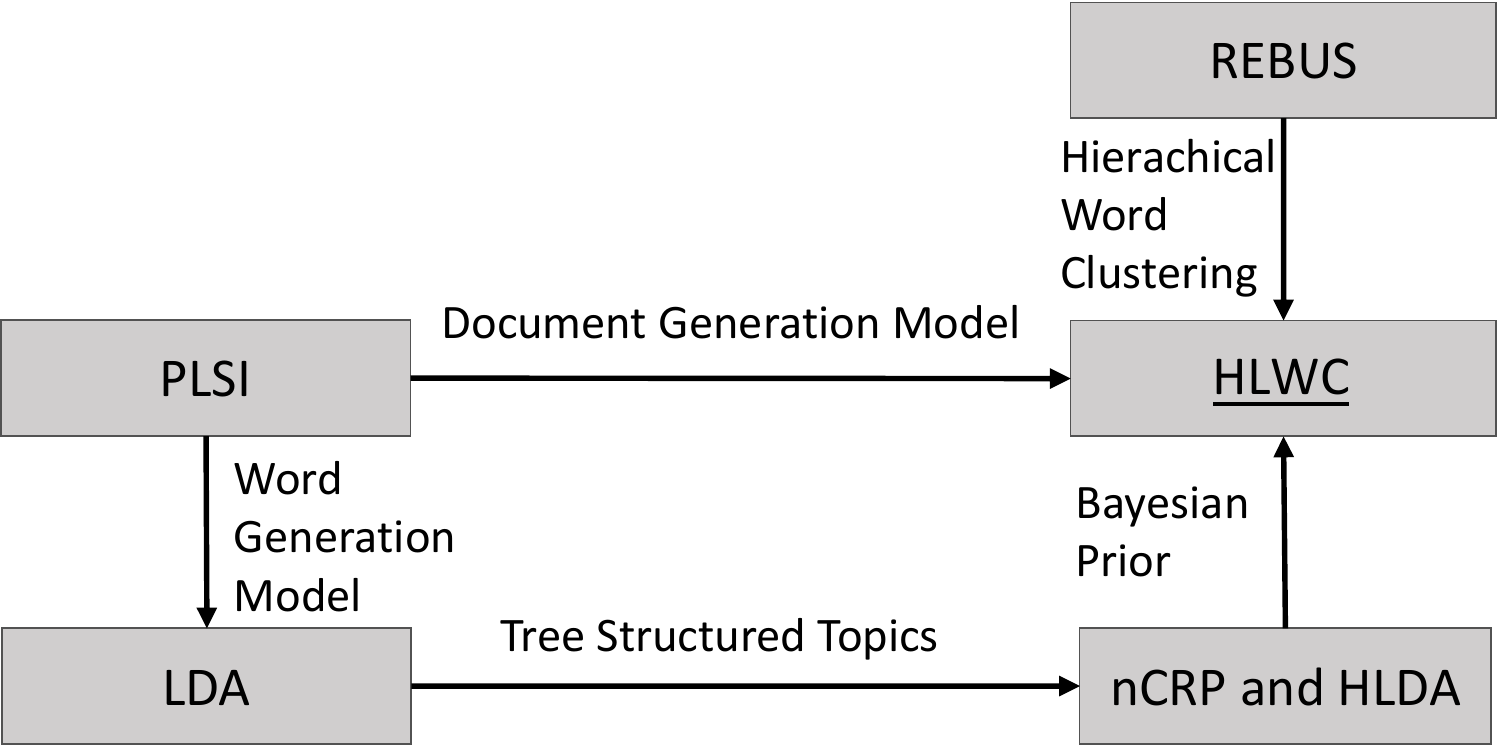}\protect
\par\end{centering}

}
\par\end{center}%
\end{minipage}%
\begin{minipage}[t]{0.5\columnwidth}%
\begin{center}
\subfloat[nCRP prior\label{fig:nCRP-prior}]{\protect\begin{centering}
\protect\includegraphics[scale=0.25]{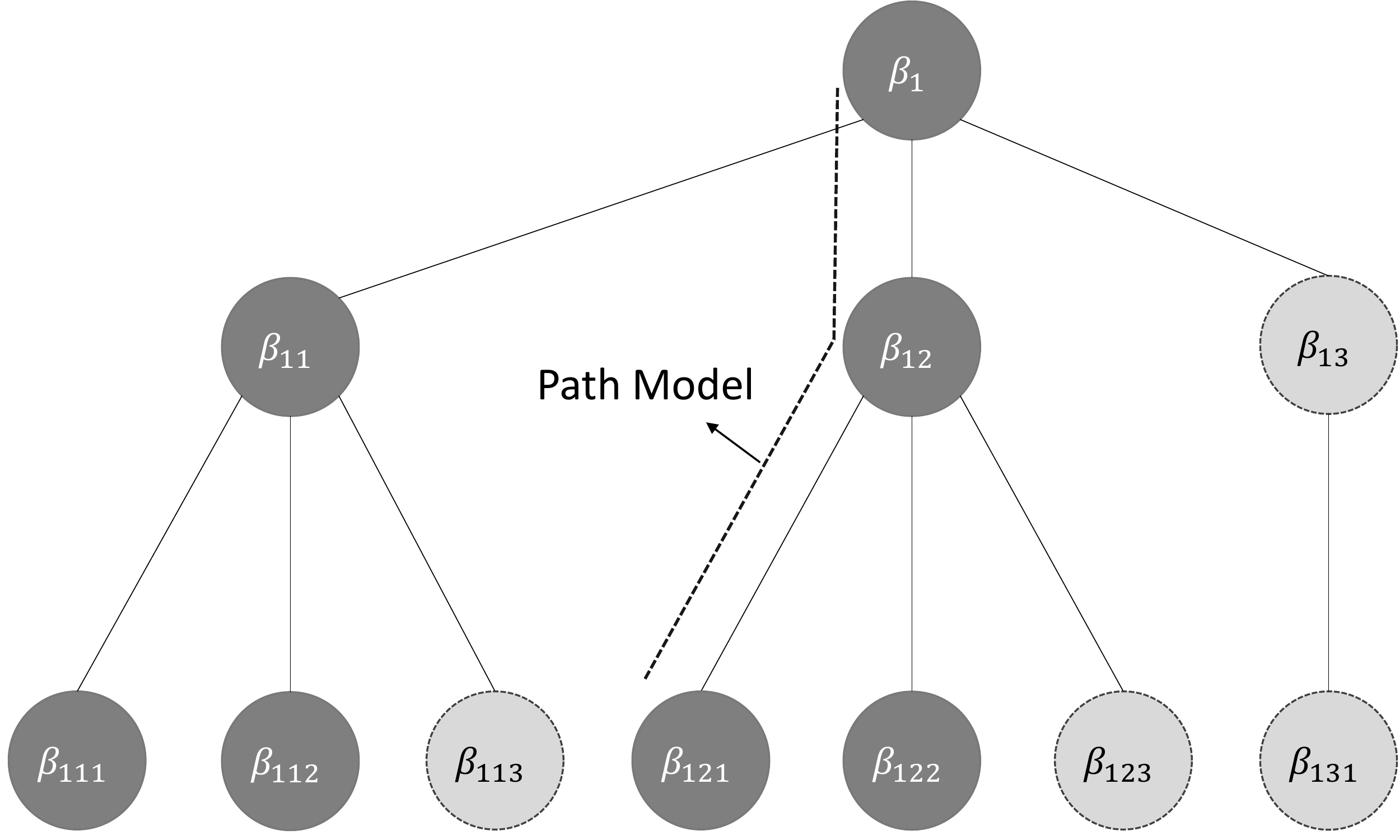}\protect
\par\end{centering}

} 
\par\end{center}%
\end{minipage}
\par\end{centering}
\vspace{0.20cm}

\begin{centering}
\begin{minipage}[t]{0.38\columnwidth}%
\begin{center}
\subfloat[HLDA Path Model\label{fig:HLDA-Path-Model}]{\protect\begin{centering}
\protect\includegraphics[scale=0.175]{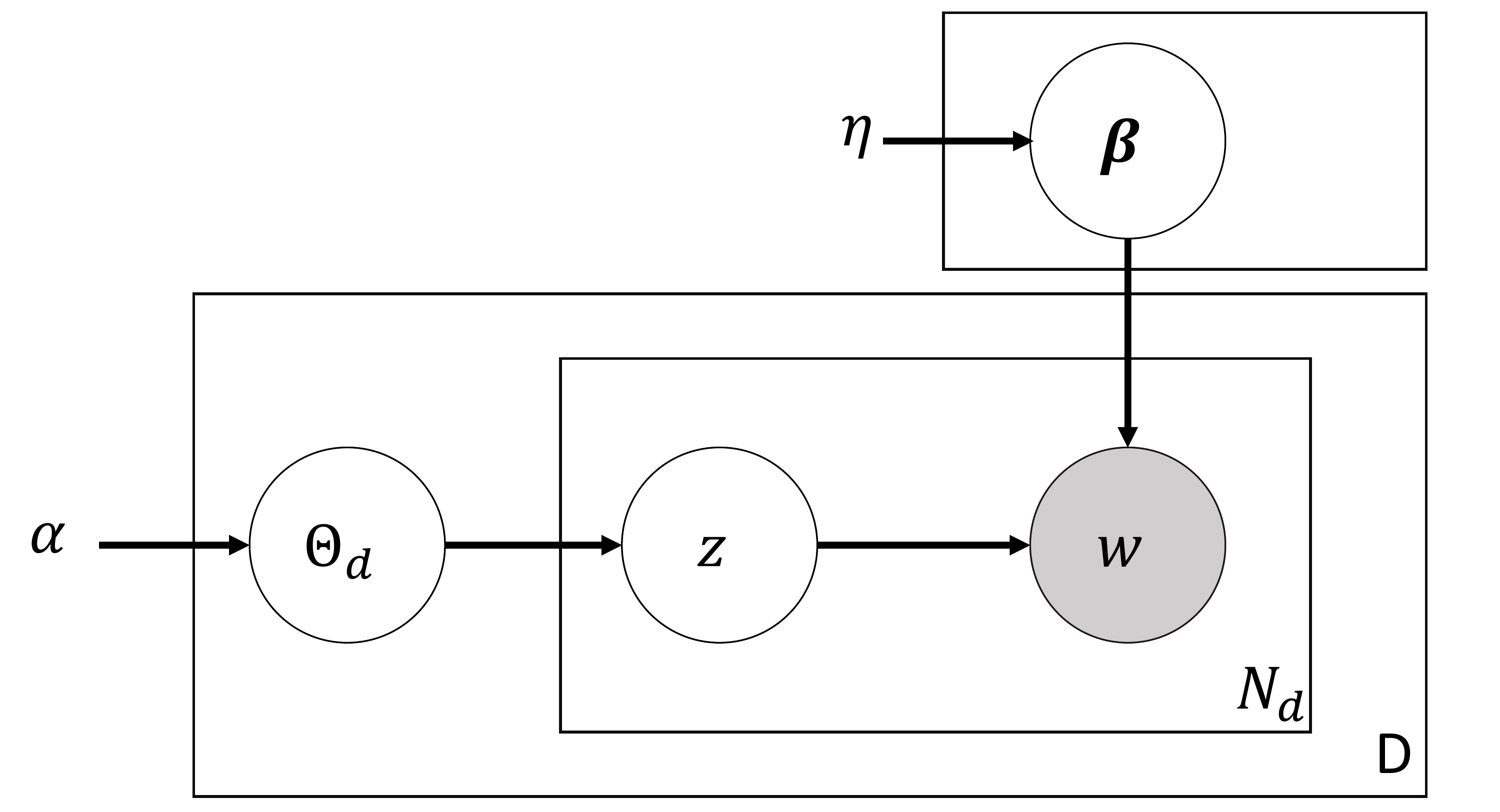}\protect
\par\end{centering}

}
\par\end{center}%
\end{minipage}%
\begin{minipage}[t]{0.34\columnwidth}%
\begin{center}
\subfloat[HLWC Path Model\label{fig:HLWC-Path-Model}]{\protect\begin{centering}
\protect\includegraphics[scale=0.175]{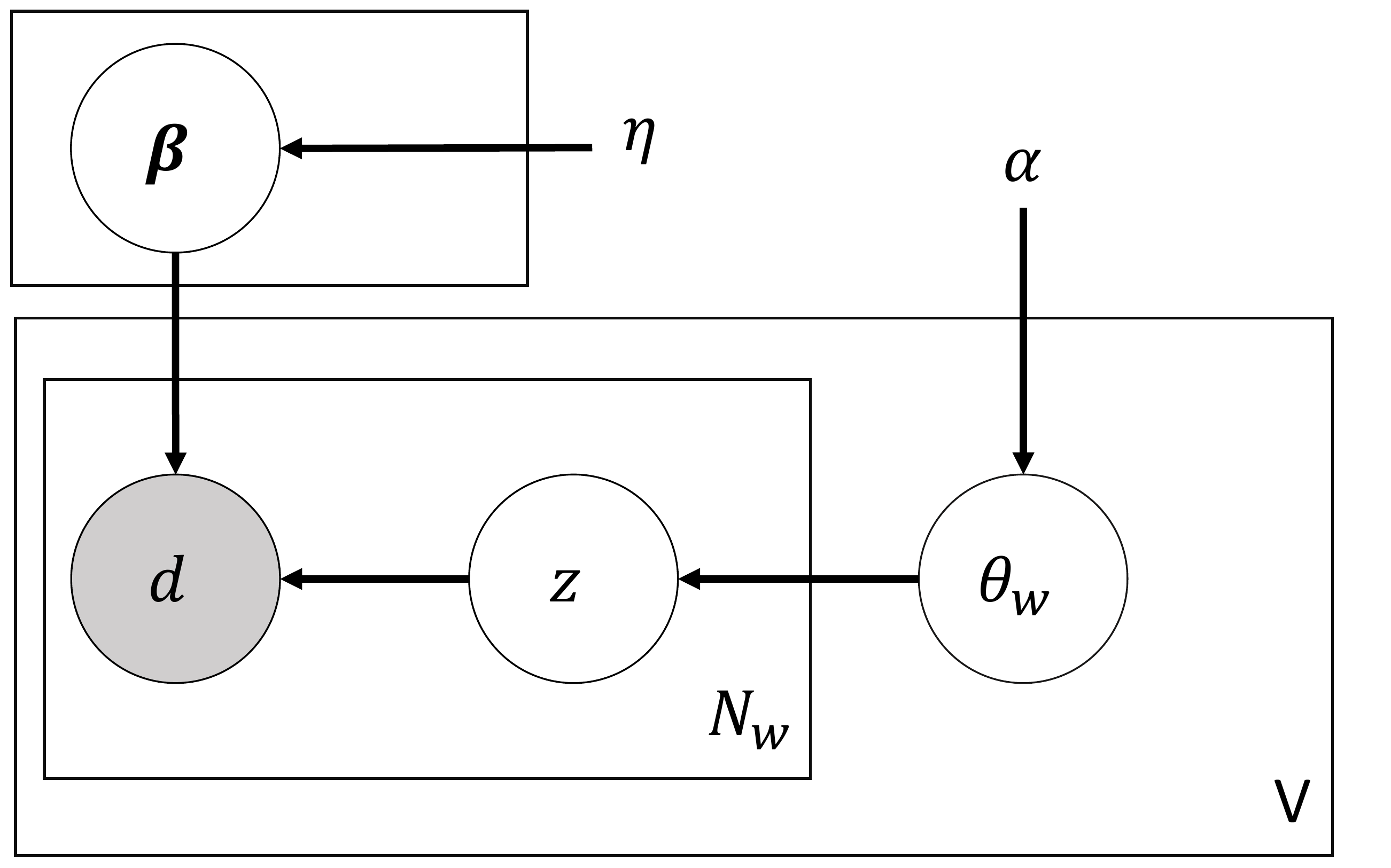}\protect
\par\end{centering}
}
\par\end{center}%
\end{minipage}
\begin{minipage}[t]{0.24\columnwidth}%
\textbf{Variables:}\\
$\beta$: Topic\\
$\theta$: Mixture weights\\
$\eta, \alpha$: Dirichlet params\\
$d$: document id \\
$z$: level allocation\\
$w$: word\\
$N_w$: number of word observations\\
\end{minipage}
\par\end{centering}

\protect\caption{nCRP prior, HLWC relations and path models}
\end{figure}

Specifically, we modified the generative model of HLDA using the document generation
model of PLSI \cite{hofmann1999probabilistic}. Whereas, topic model studies followed the word generation branch of PLSI. Our model shares the same non-parametric prior, nested Chinese Restaurant Process (nCRP), with HLDA. However, it produces hierarchical non-parametric word clusters like REBUS \cite{fidaner2014clustering} instead of document clusters as in HLDA.
Relations with existing methods are illustrated in Figure \ref{fig:HLC-model-relations}.

In the following section, we describe our model and inference. We have
given experimental results on section \ref{sec:Results}. The last section
concludes the text and gives future directions. 

\section{Hierarchical Latent Word Clustering}

nCRP prior used in HLWC was first presented in \cite{Blei04hierarchicaltopic}. 
nCRP is an exchangeable distribution on 
tree structures with unbounded depth and branching factor.   
A sample tree structure is shown in Figure \ref{fig:nCRP-prior}.
The observed nodes are shown in dark color. Light nodes are potential new branches
providing non-parametric behavior for incoming data. The path model is the model that describes the conditional distribution of data given the path. Parameters associated with each node are used in path model to generate data. 
The difference between HLDA and HLWC comes in the path model. In each path of HLDA,
documents follow LDA distribution as shown in Figure \ref{fig:HLDA-Path-Model}.
Each document has its own distribution across levels
$\theta_{d}$ and level allocation variable $z_{di}$ is sampled from that
distribution. Then, the word  $w_{di}$ is sampled from selected multinomial topic
$\beta_{z_{di}}$. In this notation, $D$ is used for a number of documents
and $N_{d}$ is used for the number of words in that document. Although it
is not explicitly shown, the document ids
along with words are observed variables of the model. 

We modified path model using document generation model of PLSI \cite{hofmann1999probabilistic} as shown in Figure \ref{fig:HLWC-Path-Model}.
The generative model for HLWC is according to Equation (\ref{eq:hlwc}). In this model,
each word $w$ chooses its path $c_w$ according to nCRP prior. Then, document ids are generated from the path model. In other words, each word generates distribution over levels $\theta_w$ from a Dirichlet distribution parameterized with $\alpha$. Then, level allocations  for word observation $z_{wi}$ are sampled from $\theta_{w}$ and document ids $d_{wi}$ are sampled from multinomial distributions $\beta_{z_{wi}}$. Topic $\beta_i$'s are distributed i.i.d. according to Dirichlet prior with parameter $\eta$. In HLWC $N_w$ represents total number of observation of a word in the whole corpus. Also, $N_d$ represents number of documents in the corpus.

\begin{eqnarray}
c_{w} & \sim & nCRP(\gamma) \nonumber\\
\theta_{w} & \sim & Dirichlet(\alpha) \nonumber\\
\beta_{i} & \sim & Dirichlet(\eta) \label{eq:hlwc}\\
z_{wi} & \sim & Multinomial(\theta_{w}) \nonumber\\
d_{wi} & \sim & Multinomial(\beta_{z_{wi}})  \nonumber
\end{eqnarray}

The parameters of the model are $\gamma,\eta, L$ and $\alpha$. $\gamma$ is nCRP prior parameter affecting the probability of new branches. Higher values of $\gamma$ tend to create wider trees. The parameter $\eta$ controls the sparseness of document distributions. For smaller values of $\eta$, model tends to create smaller clusters. Number of levels $L$ changes the depth of the tree. More levels could slow down inference and may create noisy clusters because the nodes may not have enough  data to reliably estimate the parameter. The depth is restricted to $L=3$ levels in this study. $\alpha$ controls the spareness of distribution over levels. If the level distribution is sparse words tend to belong only one level. 

The inference task is getting a representative sample tree from the posterior distribution. That is, finding level allocations of document observations and paths of the words.  
It is possible to utilize similar inference methods with HLDA since
transpose operation on dataset (changing rows with columns in bag of words representation)
creates the desired model. In a similar manner, intermediate parameters are integrated out in collapsed Gibbs sampler.
The resulting distribution of integration in a node becomes Dirichlet-multinomial compound  distribution that is denoted by DM. 
We sampled path of each word  according to the distribution (\ref{eq:path}). Then, for each observation of word in a document,  
the level allocation is sampled according to the distribution (\ref{eq:level}). 

In our notation, we have used capital bold letters for collection of variables (i.e. $\mathbf{C}=\{c_w\}_{w=1}^{V}$, $\mathbf{D_w} = \{d_{wi}\}_{i=1}^{N_w}$, $\mathbf{Z_w} = \{z_{wi}\}_{i=1}^{N_w}$). 
Minus sign (-) in power represents the exclusion of variables which is common in collapsed Gibbs sampling. 
To represent the frequencies, we have used symbol '\#'. For example $\#[\mathbf{D_{c_{w},z_{wi}}}=d_{wi}|c_{w},z_{wi}]$ represents
how many times the document id $d_{wi}$ appears in selected topic indexed by path $c_w$ and level $z_{wi}$. 

\begin{eqnarray}
P(c_{w}|\gamma,\mathbf{C^{-w}},\mathbf{Z},\mathbf{D}) & \propto & nCRP(c_{w}|\gamma,\mathbf{C^{-w}})DM(\mathbf{D_{w}}|\mathbf{Z},\mathbf{D^{-w}},\eta,c_{w}) \label{eq:path}\\
P(z_{wi}|\alpha,\mathbf{Z_{w}^{-wi}},c_{w},d_{wi}) &  \propto & (\#[\mathbf{Z_{w}^{-wi}}=z_{wi}|c_{w}]+\alpha)*\frac{\#[\mathbf{D_{c_{w},z_{wi}}}=d_{wi}|c_{w},z_{wi}]+\eta}{\#[\mathbf{D_{c_{w},z_{wi}}}|c_{w},z_{wi}]+N_d\eta} \label{eq:level}
\end{eqnarray}

\section{Results\label{sec:Results}}

We performed our inference on two datasets. First one
is NIPS dataset, a widely used text corpus used in topic modeling literature.
Also, we have collected radiology reports from publicly available
data sources. The collection included IDash radiology reports, OpenI
chest x-ray reports and Medical NLP Challenge dataset \footnote{http://www-users.cs.umn.edu/~bthomson/medicalchallenge/index.html}. 

In NIPS dataset, top 50 most frequent words were removed. We chose following
4k words for clustering without applying stemming on the vocabulary. We restricted 
number of levels to 3 in nCRP prior. Parameters of the model was $\eta=[1,1,1]$
, $\gamma=1$ and $\alpha=[1,1,1]$ for NIPS dataset. This setting corresponds to less informative prior. Collapsed Gibbs
sampler ran for 2.5k iterations. Part of the resulting hierarchy is shown
in Figure  \ref{fig:NIPS}. Top node was not drawn since it is shared by all words. 
Rest of the tree can be found in the supplementary
material. It could be seen that the model is able to group related words in the same cluster. 
Thanks to common distribution on documents, related word clusters were combined in the upper level.

\begin{figure}
\begin{centering}
\includegraphics[scale=1.12]{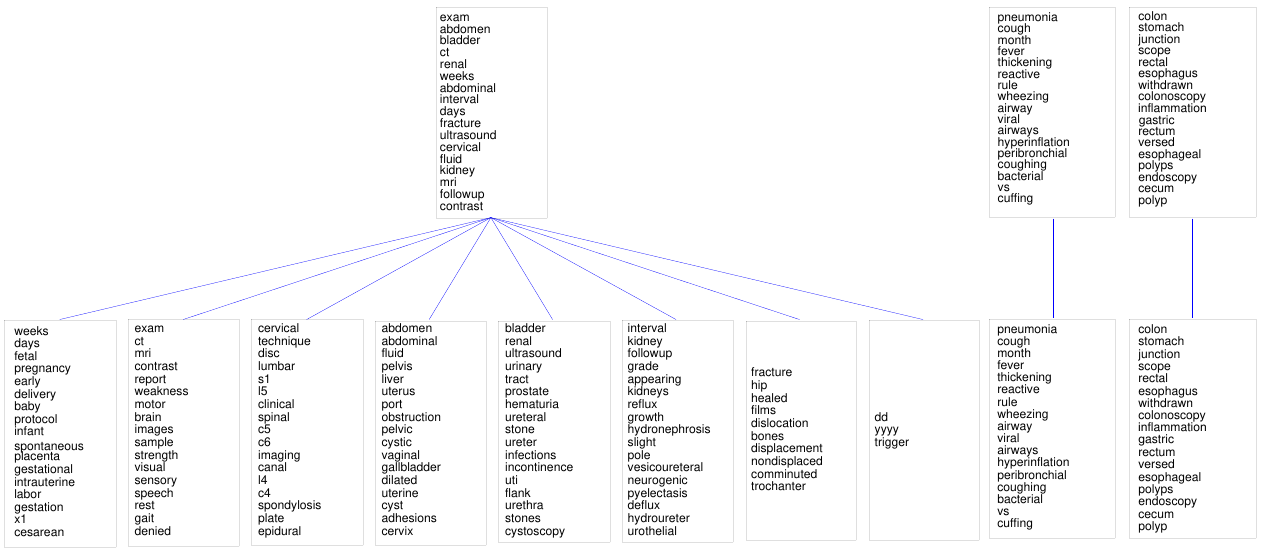}
\par\end{centering}

\protect\caption{Part of the hierarchy from radiology reports \label{fig:Part-of-hierarchy} }

\end{figure}

We collected 8452 documents to conduct experiments on radiology reports. Vocabulary size for this corpus was 5046.
We used parameters $\eta=[1,1,1]$ , $\gamma=1$ and $\alpha=[0.5,0.5,0.5]$.
After 5000 iterations, we obtained a tree structure where part of the hierarchy is
given in Figure \ref{fig:Part-of-hierarchy}. Similarly, we displayed after the second
level. In our results, imaging studies were collected in a high-level cluster. At the third level, we observed more 
specific clusters for different body locations. Some of the nodes in the second level did not create additional 
branches in the third level.

\section{Conclusion}

We presented Hierarchical Latent Word Clustering as a non-parametric hierarchical
clustering structure on words. Proposed algorithm defines topics as
multinomial distributions over documents and words those are sharing similar
document distributions are clustered in a tree. We conducted experiments
on two real-world datasets: NIPS and radiology reports. Results indicate
that word clusters could be identified with proposed inference algorithm. 
This study suggests a potentially fruitful new direction in text analysis. It is possible to
extend this study with more informative features like Word2Vec skip-gram features. 
Also, it could be possible to obtain more coherent clusters with allowing some degree of polysemy.

\subsubsection*{Acknowledgments}

We would like to thank Ferit Akova for insightful suggestions.

\begin{figure}
\begin{centering}
\includegraphics[scale=1.12]{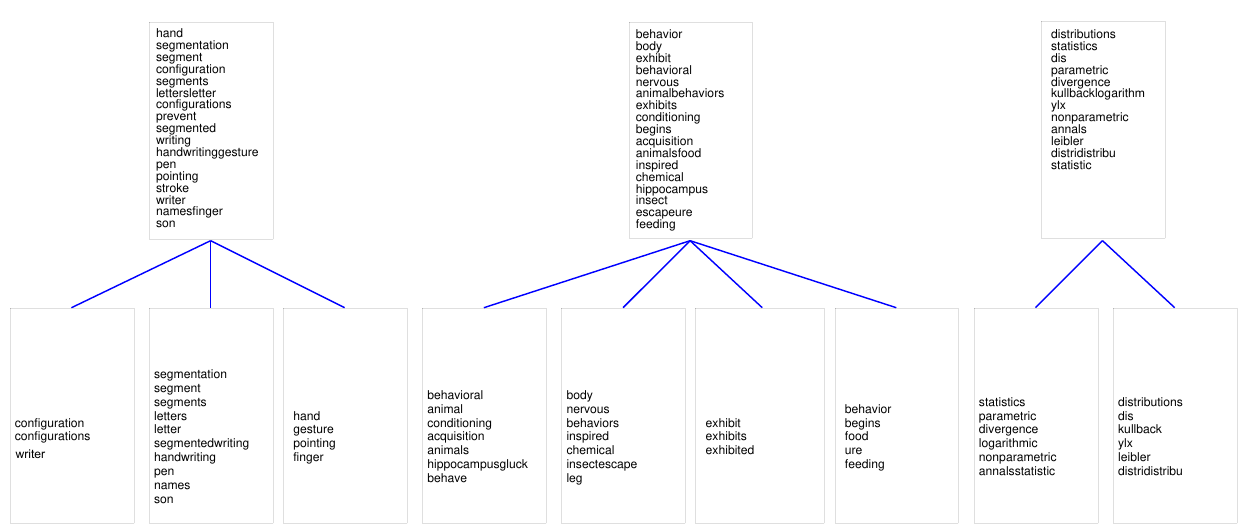}
\par\end{centering}

\protect\caption{Part of the hierarchy extracted from NIPS corpus \label{fig:NIPS}}

\end{figure}

\bibliographystyle{plain}
\bibliography{hlc}

\end{document}